\documentclass[runningheads,a4paper]{llncs}

\usepackage{amssymb}
\usepackage{amsmath}
\usepackage{graphicx}
\usepackage{cite}
\usepackage{url}
\urldef{\mailsa}\url{stephan.baier@campus.lmu.de}
\urldef{\mailsb}\url{yunpu.ma@siemens.com}
\urldef{\mailsc}\url{volker.tresp@siemens.com}

\usepackage{subcaption}
\begin{document}

\mainmatter  % start of an individual contribution

\title{Improving Visual Relationship Detection using Semantic Modeling of Scene Descriptions}

\titlerunning{Visual Relationship Detection using Semantic Modeling}

\author{Stephan Baier\textsuperscript{1}
\and Yunpu Ma\textsuperscript{1,2}
\and Volker Tresp\textsuperscript{1,2}}
\authorrunning{Stephan Baier, Yunpu Ma, Volker Tresp}

\institute{\textsuperscript{1} Ludwig Maximilian University, 80538 Munich, Germany \\
\mailsa\\
\textsuperscript{2} Siemens AG, Corporate Technology, Munich, Germany \\
\mailsb\\
\mailsc\\
}


%\toctitle{Lecture Notes in Computer Science}
%\tocauthor{Authors' Instructions}
\maketitle

\begin{abstract}

Structured scene descriptions of images are useful for the automatic processing and querying of large image databases. We show how the combination of a semantic and a visual statistical model can improve on the task of mapping images to their associated scene description. In this paper we consider scene descriptions which are represented as a set of triples (\textit{subject}, \textit{predicate}, \textit{object}), where  each triple consists of a pair of visual objects, which appear in the image, and the relationship between them (e.g. \textit{man-riding-elephant}, \textit{man-wearing-hat}). We combine a standard  visual model for object detection, based on convolutional neural networks, with a latent variable model for link prediction. We apply multiple state-of-the-art link prediction methods and compare their capability for visual relationship detection. One of the main advantages of link prediction methods is that they can also generalize to triples, which have never been observed in the training data. Our experimental results on the recently published Stanford Visual Relationship dataset,  a challenging real world dataset, show that the integration of a semantic model using link prediction methods can significantly improve the results for visual relationship detection. Our combined approach achieves superior performance compared to the state-of-the-art method from the Stanford computer vision group.

\keywords{Visual Relationship Detection, Knowledge Graph, Link Prediction}
\end{abstract}

\section{Introduction}
\label{section_introduction}

\begin{figure}[t]

\centering
\includegraphics[width=\textwidth]{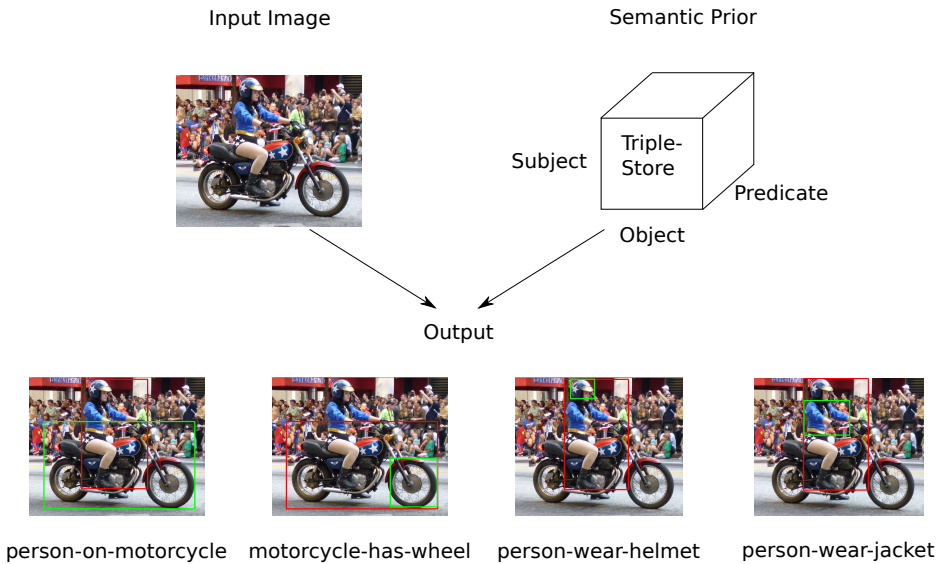}
\caption{The input to the model is a raw image. Together with a semantic prior we generate triples, which describe the scene.}
\label{fig_example}
\end{figure}

Extracting semantic information from unstructured data, such as images or text, is a key challenge in artificial intelligence. Semantic knowledge in a machine-readable form is crucial for many applications such as search, semantic querying and question answering.

Novel computer vision algorithms, mostly based on convolutional neural networks (CNN), have enormously advanced over the last years. Standard applications are  image classification and, more recently, also the detection of objects in images. However, the semantic expressiveness of describing an image simply by a set of objects is rather limited.  Semantics is captured in much more meaningful ways in the relationships between objects. In particular, visual relationships can be represented by triples of the form (\textit{subject}, \textit{predicate}, \textit{object}), where two entities appearing in an image are linked trough a relation (e.g. \textit{man-riding-elephant}, \textit{man-wearing-hat}). 

Extracting triples, i.e. visual relationships, from raw images is a challenging task, which has been a focus in the Semantic Web community \cite{sw1, sw2, sw3, sw4} but only recently has gained substantial attention in main stream computer vision \cite{phrase1, phrase2, visual}. First approaches used a single classifier, which takes an image as input and outputs an entire triple \cite{phrase1, phrase2}. However, these approaches do not scale to datasets with many object types and relationships, due to the exploding combinatorial complexity. Recently, \cite{visual} proposed a method, which classifies the visual objects and their relationships in independent preprocessing steps, and then derives a prediction score for the entire triple. This approach was applied  to the extraction of triples from a large number of possibilities. In the same paper, the first large-scale dataset for visual relationship extraction was published.

The statistical modeling of graph-structured knowledge bases, often referred to as knowledge graphs, has recently gained growing interest. The most popular approaches learn embeddings for the entities and relations in the knowledge graph. Based on the embeddings a likelihood for a triple can be derived. This approach has mainly been used for link prediction, which tries to predict missing triples in a knowledge graph. A recent review paper can be found in \cite{relational_review}.

In the approach described in this paper, statistical knowledge base models are used to support  the task of visual link prediction. Note that in many cases a high likelihood for a triple can be inferred using a semantic model. For example if the visual model detects a motorbike, it is very likely that the triple \textit{motorbike-has\_part-wheel} is true, as all motorbikes have a wheel. We suggest that integrating such prior knowledge can improve various computer vision tasks. In particular,  we propose to combine the likelihood from a semantic model and a visual model to enhance the prediction of image triples. 

Figure \ref{fig_example} illustrates our approach. The model takes as input a raw image and adds a semantic prior, which is derived from the training data. Both types of information are combined, to predict the output, which consists of a set of triples describing the scene, and their corresponding bounding boxes.

For combining the semantic prior with the visual model, we present a probabilistic view on the problem which can be divided into a semantic part and a visual part. We show how the semantic part of the probabilistic model can be implemented using standard link prediction methods and the visual part using recent computer vision algorithms.

We train our semantic model by using absolute frequencies from the training data, describing how often a triple appears in the training data. By applying a latent variable model, we are able to also generalize to unseen or rarely seen triples, which still have a high likelihood of being true due to their similarity to other likely triples. For example if we frequently observe the triple \textit{person-ride-motorcycle} in the training data we can generalize also to a high likelihood for \textit{person-ride-bike} due to the similarity between \textit{motorcycle} and \textit{bike}, even if the triple \textit{person-ride-bike} has not been observed or just rarely been observed in the training data. The similarity of \textit{motorcycle} and \textit{bike} can be derived from other triples, which describe for example that both have a \textit{wheel} and both have a \textit{handlebar}.

We conduct experiments on the Stanford Visual Relationship dataset recently published by \cite{visual}. We evaluate different model variants on the task of predicting semantic triples and the corresponding bounding boxes of the subject and object entities detected in the image. Our experiments show, that including the semantic model improves on the state-of-the-art result in the task of mapping images to their associated triples.

The paper is structured as follows. Section \ref{section_related_work} gives an overview of the state-of-the-art link prediction models, the employed computer vision techniques and related work. Section \ref{section_model} describes the semantic and the visual part of our model and how both can be combined in a probabilistic framework. In Section \ref{section_experiments} we show a number of different experiments. Finally, we conclude our work with Section \ref{section_conclusion}.

%Allgemein:
%
%- Embeddings, Representation Learning
%- Symmetric Relations, etc.

%-Related work (image captioning etc.)
%
%The problem of extracting descriptions has gained some popularity recently. Question answering systems and image captioning.
%
%-Success in modeling knowledge bases
%
%Latent variable models for modeling large scale knowledge bases have become popular.
%
%Great improvement in vision and semantic modeling

\section{Background and Related Work}
\label{section_related_work}

Our proposed model joins ideas from two areas, computer vision and statistical relational learning for semantic modeling. Both fields have developed rapidly in recent years. In this chapter we discuss relevant work from both areas.

\subsection{Statistical Link Prediction}
\label{section_related_work_link_prediction}
A number of statistical models have been proposed for modeling graph-structured knowledge bases often referred to as knowledge graphs. Most methods are designed for predicting missing links in the knowledge graph. A recent review on link prediction can be found in \cite{relational_review}. A knowledge graph $\mathcal{G}$ consists of a set of triples $\mathcal{G} = \{(s, p, o)_i\}_{i=1}^N \subseteq \mathcal{E} \times \mathcal{R} \times \mathcal{E}$. The entities $s, o \in \mathcal{E}$ are referred to as \textit{subject} and \textit{object} of the triple, and the relation between the entities $p \in \mathcal{R}$ is referred to as \textit{predicate} of the triple.

Link prediction methods can be described by a function $\theta: \mathcal{E} \times \mathcal{R} \times \mathcal{E} \rightarrow \mathbb{R}$, which maps a triple $(s,p,o)$ to a real valued score. The score of a triple $\theta(s,p,o)$ represents the likelihood of the triple being true. Most recent link prediction models learn a latent representation also called embedding for the entities and the relations. In the following we describe state-of-the-art link prediction methods, which we will use later in our approach to visual relationship detection.

\paragraph{DistMult:} DistMult \cite{DistMult} scores a triple by building the tri-linear dot product of the embeddings, such that
\begin{equation}
\label{eq_link_first}
\theta(s, p, o) = \langle a(s), r(p), a(o)\rangle = \sum_{j} a(s)_j r(p)_j a(o)_j
\end{equation}
where $a: \mathcal{E} \rightarrow \mathbb{R}^d$ maps entities to their latent vector representations and similarly $r: \mathcal{R} \rightarrow \mathbb{R}^d$ maps relations to their latent representations. The dimensionality $d$ of the embeddings, also called rank, is a hyperparameter of the model.

\paragraph{ComplEx:} ComplEx \cite{ComplEx} extends DistMult to complex valued vectors for the embeddings of both, relations and entities. The score function is
\begin{equation}
\begin{split}
\theta(s,p,o) = Re(\langle a(s), r(p), \overline{a(o)}\rangle) &= \langle Re(a(s)), Re(r(p)), Re(a(o))\rangle \\
&+ \langle Im(a(s)), Re(r(p)), Im(a(o))\rangle \\
&+ \langle Re(a(s)), Im(r(p)), Im(a(o))\rangle \\
&- \langle Im(a(s)), Im(r(p)), Re(a(o))\rangle 
\end{split}
\end{equation}
where $a: \mathcal{E} \rightarrow \mathbb{C}^d$ and $r: \mathcal{R} \rightarrow \mathbb{C}^d$; $Re(\cdot)$ and $Im(\cdot)$ denote the real and imaginary part, respectively, and $\overline{\cdot}$ denotes the complex conjugate.

\paragraph{Multiway NN:} The multiway neural network \cite{knowledge_vault, relational_review} concatenates all embeddings and feeds them to a neural network of the form
\begin{equation}
\theta(s,p,o) = \left( \beta^T \tanh \left( A \left[a(s), r(p), a(o) \right] \right) \right)
\end{equation}
where $\left[\cdot, \cdot, \cdot \right] $ denotes the concatenation of the embeddings $a(s), r(p), a(o) \in \mathbb{R}^d$. $A$ is a weight matrix and $\beta$ a weight vector.

\paragraph{RESCAL:} The tensor decomposition RESCAL \cite{Rescal} learns vector embeddings for entities and matrix embeddings for relations. The score function is
\begin{equation}
\label{eq_link_last}
\theta(s,p,o) = a(s) \cdot R(p) \cdot a(o)
\end{equation}
with $\cdot$ denoting the dot product, $a: \mathcal{E} \rightarrow \mathbb{R}^d$ and $R: \mathcal{R} \rightarrow \mathbb{R}^{d \times d}$.

Typically, the models are trained using a ranking cost function \cite{relational_review}. For our task of visual relationship detection, we will train them slightly differently, as we show in section \ref{section_model_semantic}. Another popular link prediction method is TransE \cite{transe}, however it gives negative scores, which is not appropriate for modeling count data as we will see in the next section; thus we are not considering it in this work.

\subsection{Image Classification and Object Detection}

Computer vision methods for image classification and object detection have improved enormously over the last years. Convolutional neural networks (CNN), which apply convolutional filters in a hierarchical manner to an image, have become the standard for image classification. In this work we use the following two methods.

\paragraph{VGG:} The VGG-network is a convolutional neural network, which has shown state-of-the-art performance at the Imagenet challenge \cite{vgg}. It exists in two versions, i.e. the VGG-16 with 16 convolutional layers and VGG-19 with 19 convolutional layers.

\paragraph{RCNN:} The region convolutional neural network (RCNN) \cite{rcnn} proposes regions, which show some visual objects in the image. It uses a selective search algorithm for getting candidate regions in an image \cite{selective_search}. The RCNN algorithm then rejects most of the regions based on a classification score. As a result, a small set of region proposals is derived. There are two extentions to RCNN, which are mainly faster and slighly more accurate \cite{fast_rcnn} \cite{faster_rcnn}. However, in our experiments we use the original RCNN, for a fair comparison with \cite{visual}. Our focus is on improving visual relationship detection trough semantic modeling rather than on improving computer vision techniques.

\subsection{Visual Relationship Detection}

Visual relationship detection is about predicting triples from images, where the triples consist of two visual objects and the relationship between them. This is related to visual caption generation, which gained considerable popularity among the deep learning community recently, where an image caption consisting of natural text is generated given an image  \cite{caption1, caption2}. However, the output in visual relationship detection is more structured (a set of triples), and thus it is more appropriate for further processing, e.g. semantic querying. The extraction of semantic triples has been successfully applied to text documents, e.g. the Google Knowledge Vault project for improving the Google Knowledge Graph \cite{knowledge_vault}.

Some earlier work on visual relationship detection was concerned with learning spatial relationships between objects, however with a very limited set of only four spatial relations \cite{spatial1, spatial2}. Other related work attempted to learn actions and object interactions of humans in videos and images \cite{action1,action2,action3,action4,action5}. Full visual relationship detection has been demonstrated in \cite{phrase1, phrase2}, however, also with only small amounts of triples. 

The Stanford computer vision group was first in proposing a scalable model and applying it to a  large-scale dataset, with 700000 possible triples. In their work,  entities of the triples were  detected separately and a joint score for each triple candidate was computed \cite{visual}. The visual module in \cite{visual} uses the following computer vision methods, which we will also use in our approach. A RCNN for object detection is used to derive candidate regions. Further, a VGG-16 is applied to the detected regions for obtaining object classification scores for each region. Finally, a second VGG, which classifies relationships, such as \textit{taller-than}, \textit{wears}, etc. is applied to the union of pairs of regions.  The model also contains a language prior, which can model semantic relationships to some extend based on word embeddings. The language prior allows the model to generalize to unseen triples. However, our experiments show that integrating state-of-the-art link prediction methods for modeling semantics is more appropriate for improving general prediction and generalization to unseen triples.

\section{Modeling Visual Relationships}
\label{section_model}

\begin{figure}[t]
\centering
\includegraphics[width=0.4\textwidth]{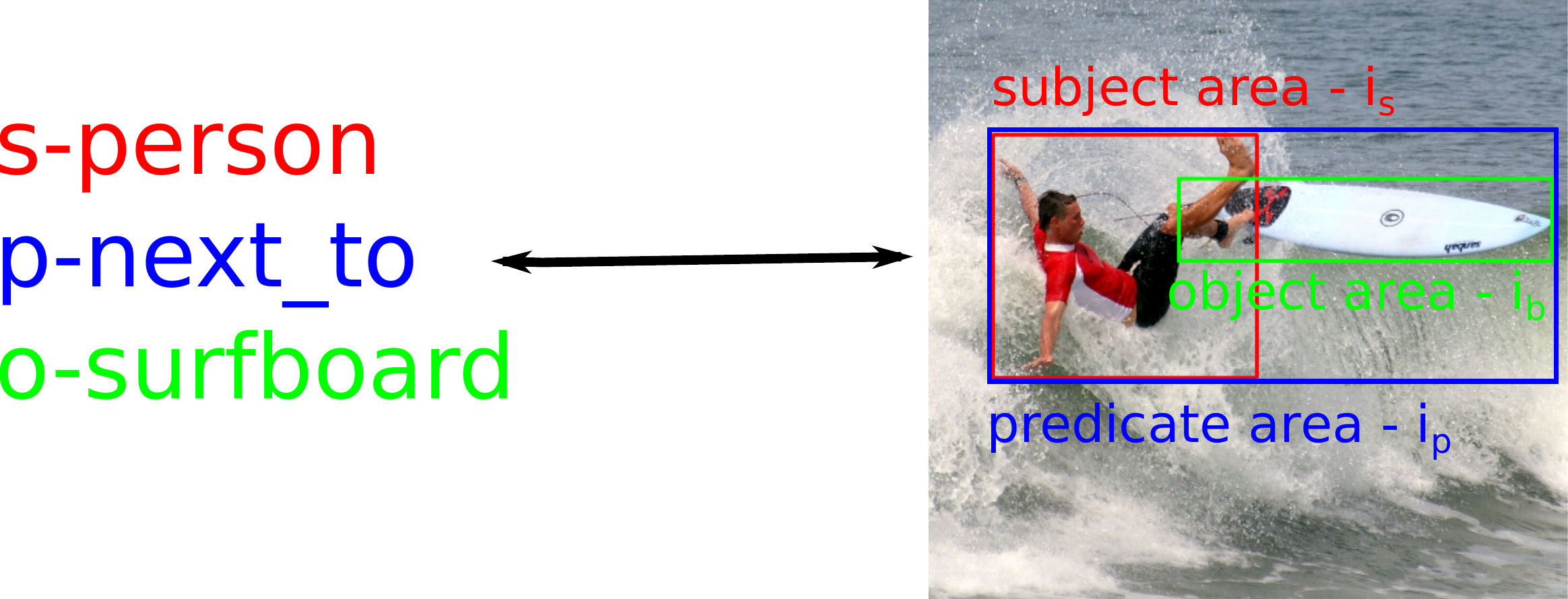}
\caption{The subject and object of the triple relate to two regions in the image, and the predicate relates to the union of the two regions.}
\label{fig_areas}
\end{figure}

In the following we describe our approach to jointly modeling images and their corresponding semantics. 

\subsection{Problem Description}

\begin{figure}[t]

\centering
\includegraphics[width=\textwidth]{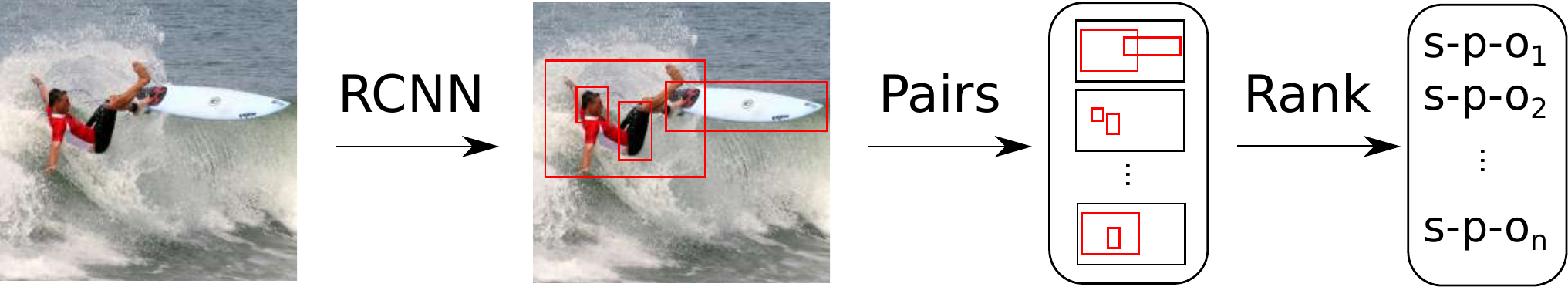}
\caption{The pipeline for deriving a ranked list of triples is as follows: The image is passed to a RCNN, which generates region candidates. We build pairs of regions and predict a score for every triple, based on our ranking method. The visual part is similar to \cite{visual}, however the ranking method is different as it includes a semantic model.}
\label{fig_pipeline}
\end{figure}

We assume data consisting of images and corresponding triple sets. For each \textit{subject} $s$ of a triple $(s,p,o)$ there exists a corresponding region $i_s$ in the image. Similarly, each \textit{object} $o$ corresponds to an region $i_o$, and each \textit{predicate} $p$ to an region $i_p$, which is the union of the regions $i_s$ and $i_o$. Thus, one data sample can be represented as a six-tuple of the form $(i_s, i_p, i_o, s, p, o)$. Figure \ref{fig_areas} shows an example of a triple and its corresponding bounding boxes. During training all triples and their corresponding areas are observed. After model training the task is to predict the most likely tuples $(i_s, i_p, i_o, s, p, o)$ for a given image. Figure \ref{fig_pipeline} shows the processing pipeline of our method, which takes a raw image as an input, and outputs a ranked list of triples and bounding boxes.

%We denote the corresponding random variables as $I_S, I_P, I_O, S, P, O$.

%Figure \ref{fig_motivation} shows an example image in the first row. The goal is to infer the triples and their corresponding bounding boxes, as shown in the second row.
 
%As in \cite{visual} we assume that $I_{A,B}$ contains the visual information for the predicate of a triple. Figure \ref{bounding_boxes_b} shows another example, and the corresponding areas.

%The visual relationships are described by triples (s, p, o), where each s and o corresponds to an area in the image ($I_A, I_B$) and p corresponds to the joint area ($I_{AB}$).

\subsection{Semantic Model}
\label{section_model_semantic}

%In contrast to typical knowledge graph modeling, we do not only have one global graph $\mathcal{G}$, but an instance of a knowledge graph $\mathcal{G}_i$ for every image $i$. The link prediction model shall reflect the likelihood of a triple to appear in a graph instance, as a prior without seeing the image. Thus, for training the link prediction model we consider the absolute frequency of triples $(s,p,o)$ in the training data which we denote as $y_{s,p,o}$. As we are dealing with count data, we assume a Poisson distribution on the model output $\theta(s,p,o)$. The log-likelihood for a triple is

In contrast to typical knowledge graph modeling, we do not only have one global graph $\mathcal{G}$, but an instance of a knowledge graph $\mathcal{G}_i$ for every image $i$. Each triple which appears in a certain image can be described as a tuple $(s, p, o, i)$. The link prediction model shall reflect the likelihood of a triple to appear in a graph instance, as a prior without seeing the image. Thus, we marginalize out the dimension $i$ and derive the absolute frequency of the triple $(s,p,o)$ in the training data, which we denote as $y_{s,p,o}$. We aim to model $y_{s,p,o}$ using the link prediction methods described in Section \ref{section_related_work_link_prediction}. As we are dealing with count data, we assume a Poisson distribution on the model output $\theta(s,p,o)$. The log-likelihood for a triple is
\begin{equation}
\log p(y_{s,p,o} | (s,p,o), \Theta) = y_{s,p,o} \log \eta(\theta(s,p,o)) - \eta(\theta(s,p,o)) - \log(y_{s,p,o}!),
\end{equation}
where $\Theta$ are the model parameters of the link prediction method and $\eta$ is the parameter for the Poisson distribution, namely
\begin{equation}
\eta(\theta(s,p,o)) = \exp(\theta(s,p,o)). 
\end{equation}
We train the model by minimizing the negative log-likelihood. In the objective function the last term $\log(y_{s,p,o}!)$ can be neglected, as it does not depend on the model parameters. Thus the cost function for the whole training dataset becomes
\begin{equation}
\textit{cost} = \sum_{(s,p,o)} \eta(\theta(s,p,o)) - y_{s,p,o} \log \eta(\theta(s,p,o)).
\end{equation}
Using this framework, we can train any of the link prediction methods described in Section \ref{section_related_work}, by plugging it into the cost function and minimizing the cost function using a gradient-descent based optimization algorithm. In this work we use Adam, a recently proposed optimization method with adaptive learning rate \cite{adam}.

%-Latent Variable models for including prior Semantic Knowledge

%-Semantic modeling using latent variables has been successful
%-Best performing link-prediction models
%-Why does it make sense to include prior semantic knowledge + Example
%-Fusion model of both visual and semantic model
%-separate cost functions (how to motivate???)

%For modeling the semantic part ($\tilde{P}(S,P,O)$) we build a tensor $\mathcal{X} \in R^{S_N, P_N, O_N}$. A element in $X_{s, p, o}$ describes the number of occurences of this triple in the training data. The resulting tensor is very sparse, and some triples, did not occur or just rarely occurred in the training data, although they are semantically very similar to triples which have occurred frequently. In order to generalize to these triples and rate them higher, we propose to apply another latent variable model on the count tensor $X$. 

\subsection{Visual Model}
Our visual model is similar to the approach used in \cite{visual}. Figure \ref{fig_pipeline} shows the involved steps. An image is first fed to a RCNN, which generates region proposals for a given image. The region proposals are represented as bounding boxes within the image. The visual model further consists of two convolutional neural networks (CNNs). The first CNN which we denote as $\textit{CNN}_e$ takes as input the subregion of the image defined by a bounding box and classifies entities from the set $\mathcal{E}$. 

The second CNN, which we denote as $\textit{CNN}_r$ takes the union region of two bounding boxes as an input, and classifies the relationship from the set $\mathcal{R}$. While training, both CNNs use the regions (bounding boxes) provided in the training data. 

For new images, we derive the regions from the RCNN. We build all possible pairs of regions, where each pair consists of a region $i_s$ and $i_o$. We apply $\textit{CNN}_e$ to the regions, to derive the classification scores $\textit{CNN}_e(s | i_s)$ and $\textit{CNN}_e(o | i_o)$. Then the union of the regions $i_s$ and $i_o$ is fed to $\textit{CNN}_r$ to derive the score $\textit{CNN}_r(p | i_p)$, where $i_p = union(i_s, i_o)$. Figure \ref{fig_areas} shows an example of the bounding boxes of the \textit{subject} and the \textit{object}, as well as the union of the bounding boxes, which relates to the \textit{predicate} of the triple.

% VGG!!!!

%Bounding Box B_1 B_2: CNN(R | union(O_1, O_2)) CNN(E, O_1 | I) CNN(E, O_2 | I)

%\subsection{Model Training}
%
%The joint cost function factors into four parts???
%
%\subsection{Model Testing}???

\subsection{Probabilistic Joint Model}

%\begin{figure}[t]
%\begin{subfigure}[c]{0.5\textwidth}
%\centering
%\includegraphics[width=0.8\textwidth]{prob.pdf}
%\subcaption{Probabilistic graphical model.}
%\label{fig_probmodel}
%\end{subfigure}
%\begin{subfigure}[c]{0.5\textwidth}
%\centering
%\includegraphics[width=0.6\textwidth]{areas.pdf}
%\subcaption{Triples and the corresponding regions.}
%\label{fig_areas}
%\end{subfigure}
%\end{figure}

\begin{figure}[t]
\centering
\includegraphics[width=0.3\textwidth]{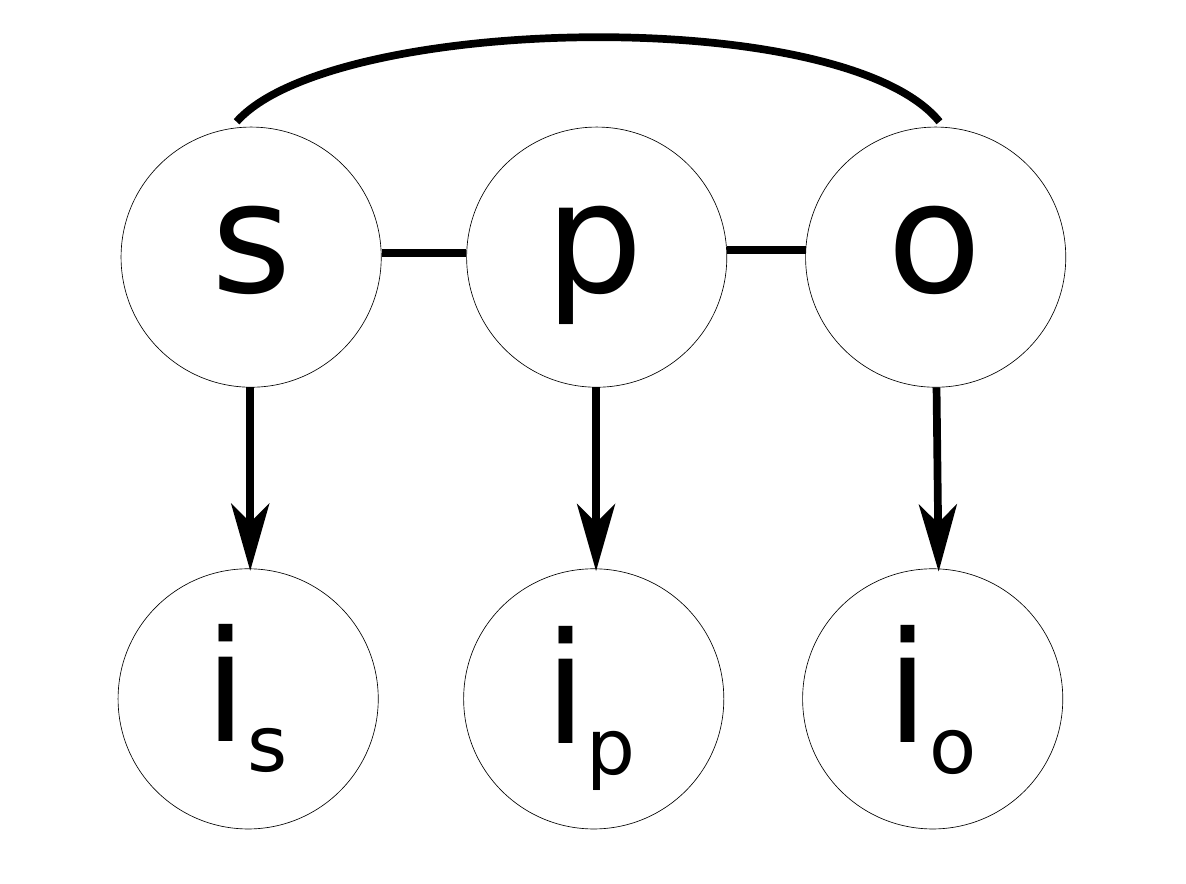}
\caption{The probabilistic graphical model describes the interaction between the visual and the semantic part for a given image. We assume the image regions $i_s$, $i_p$ and $i_o$ to be given by the RCNN.}
\label{fig_probmodel}
\end{figure}

In the last step of the pipeline in Figure \ref{fig_pipeline}, which we denote as ranking step, we need to combine the scores from the visual model with the scores from the semantic model. For joining both, we propose a probabilistic model for the interaction between the visual and the semantic part. Figure \ref{fig_probmodel} visualizes the joint model for all variables in a probabilistic graphical model. The joint distribution factors as 
 \begin{equation}
p(s, p, o, i_s, i_p, i_o) \propto \tilde{p}(s,p,o) \cdot \tilde{p}(i_s|s) \cdot \tilde{p}(i_p | p) \cdot \tilde{p}(i_o | o)
 \label{eq_joint_prob}
 \end{equation}
with $\tilde{p}$ denoting unnormalized probabilities. We can devide the joint probability of equation \ref{eq_joint_prob} into two parts. The first part is $\tilde{p}(s,p,o)$, which models semantic triples. The second part is $\tilde{p}(i_s|s) \cdot \tilde{p}(i_p | p) \cdot \tilde{p}(i_o | o)$, which models the visual part given the semantics. 

To derive the joint probability of the triples, we model $\tilde{p}(s, p, o)$ using a Boltzmann distribution such that
\begin{equation}
\tilde{p}(s, p, o) = \exp{-\beta E(s,p,o)}
\end{equation}
with $E$ being an energy function and $\beta \geq 0$ being an inverse temperature parameter. Similar to \cite{tensorbrain} we derive the energy from the link prediction model as described in Section \ref{section_model_semantic} and define it as 
\begin{equation}
E(s,p,o) = -\log \eta(\theta(s,p,o)).
\end{equation}
With a constant inverse temperature of $\beta = 1$, the unnormalized probability $\tilde{p}(s, p, o)$ becomes
\begin{equation}
\tilde{p}(s, p, o) = \eta(\theta(s,p,o)).
\end{equation}

The visual modules described in the previous section model the unnormalized probabilities $\tilde{p}(s | i_s)$, $\tilde{p}(p | i_p)$, and $\tilde{p}(o | i_o)$. By applying Bayes rule to equation \ref{eq_joint_prob} we get
\begin{equation}
p(s, p, o, i_s, i_p, i_o) \propto \tilde{p}(s,p,o) \cdot \frac{\tilde{p}(s | i_s) \cdot \tilde{p}(p | i_p) \cdot \tilde{p}(o | i_o) }{\tilde{p}(s) \cdot \tilde{p}(p) \cdot \tilde{p}(o)}.
\end{equation}
The additional terms of the denominator $\tilde{p}(s)$, $\tilde{p}(p)$, $\tilde{p}(o)$ can simply be derived from the relative frequencies in the training data.

For each image, we derive the region candidates $i_s$, $i_p$, $i_o$ from the RCNN. We do not have to normalize the probabilities as we are finally interested in a ranking of the most likely six-tuples $(i_s, i_p, i_o, s, p, o)$ for a given image. The final unnormalized probability score on which we rank the tuples is
\begin{equation}
\tilde{p}(s, p, o, i_s, i_p, i_o) =  \eta(\theta(s,p,o))  \frac{\textit{CNN}_e(s | i_s) \cdot \textit{CNN}_r(p | i_p) \cdot \textit{CNN}_e(o | i_o) }{ \tilde{p}(s) \cdot \tilde{p}(p) \cdot \tilde{p}(o)}.
\label{eq_ranking}
\end{equation}

%Each of the examples on the bottom row in figure \ref{fig_motivation}, are a sample from the distribution defined in equation \ref{joint_porb}.

%The triples s,p,o build the latent structure which induce the observed variables $I_A, I_{AB}, I_B$ in an image $I$.

\section{Experiments}
\label{section_experiments}
We evaluate our proposed method on the recently published Stanford Visual Relationship dataset \cite{visual}. We compare our proposed method against the state-of-the-art method from \cite{visual} in the task of predicting semantic triples from images. As in \cite{visual} we will devide the setting into two parts. First an evaluation on how well the methods perform when predicting all possible triples and second only evaluating on unseen triples, which is also referred to as zero-shot learning.

\subsection{Dataset}

The dataset consists of 5000 images. The semantics are described by triples, consisting of 100 entity types, such as \textit{motorcycle, person, surfboard, watch}, etc. and 70 relation types, e.g. \textit{next\_to, taller\_than, wear, on}, etc. The entities correspond to visual objects in the image. For all \textit{subject} and \textit{object} entities the corresponding regions in the image are given. Each image has in average 7.5 triples, which describe the scene. In total there are 37993 triples in the dataset. The dataset is split into 4000 training and 1000 test images. There are 1877 triples, which occur in the test set but not in the training set.

%The relationships contain actions, such as \textit{kick}, comparative statements such as \textit{taller-than}, spatial relations (e.g. \textit{next-to}) 

%\subsection{Relational Count Data}
%
%\begin{figure}[t]
%\begin{subfigure}[c]{0.4\textwidth}
%\centering
%    \includegraphics[width=\textwidth]{test-error.pdf}
%    \subcaption{Test error as a function of the rank.}
%\label{fig_test_rank}
%\end{subfigure}\hspace{0.1\textwidth}%

%
%\subsubsection{Experimental Setting}
%
%In our first experiment we evaluate the applicability of the link prediction methods to count data using the Poisson cost function. We build a dataset, which consists of the total counts of how often a triple appears in the training data. We hold out 5 percent of the non-zero triples for validating the model. 
%
%\subsubsection{Results}
%Figure \ref{fig_test_rank} shows the mean squared error on the hold out set as a function of the dimensionality of the embeddings, also called rank. Figure \ref{fig_test_iter} shows the mean squared error as a function of the iterations, with a fixed rank of 15.

\subsection{Visual Relationship Detection}
\begin{table}[t]
\centering
\caption{Results for visual relationship detection. We report Recall at 50 and 100 for four different validation settings.}
\setlength{\tabcolsep}{0.4em}
{\renewcommand{\arraystretch}{1.4}
\begin{tabular}{|c||c|c||c|c||c|c||c|c|}
\hline
 Task &\multicolumn{2}{c||}{Phrase Det.} &\multicolumn{2}{c||}{Rel. Det.} &\multicolumn{2}{c||}{Predicate Det.} & \multicolumn{2}{c|}{Triple Det.} \\
\hline Evaluation & R@100 & R@50 & R@100 & R@50 & R@100 & R@50 & R@100 & R@50\\
\hline
\hline Lu et al.  V \cite{visual} & 2.61 & 2.24 & 1.85 & 1.58 & 7.11 & 7.11 & 2.68 & 2.30 \\ 
\hline Lu et al. full \cite{visual} & 17.03 & 16.17 & 14.70 & 13.86 & 47.87 & 47.87 & 18.11 &  17.11 \\ 
%\hline Visual TransE & 22.42 & 19.42 & 15.20 & 14.07 & & \\
%\hline RESCAL ALS & 18.87 & 17.78 & 16.56 & 15.55 & 52.09  & 52.09 \\

\hline \hline RESCAL & 19.17  & 18.16  &  16.88  & 15.88  & 52.71  & 52.71 & 20.23 & 19.13  \\
\hline MultiwayNN & 18.88 & 17.75 & 16.65 & 15.57 & 51.82 & 51.82 & 19.76 & 18.53 \\ 
\hline ComplEx & 19.36 & 18.25 & 17.12 & 16.03 & 53.14 & 53.14 & 20.23 & 19.06 \\ %(rank 10)
\hline DistMult & 15.42 & 14.27 & 13.64 & 12.54 & 42.18 & 42.18 & 16.14 & 14.94 \\
\hline 
\end{tabular} 
}
\label{results_main}
\end{table}

\subsubsection{Experimental Setting}

For doing visual relationship detection, we consider four different types of settings. Three of them are identical to the experimental settings in \cite{visual}. We add a fourth setting, which eliminates the evaluation of correctly detecting the bounding boxes, and solely evaluates the predicted triples. The four settings are as follows. 

\paragraph{Phrase detection:} In phrase detection the task is to give a ranking of likely triples plus the corresponding regions for the \textit{subject} and \textit{object} of the triple. The bounding boxes are derived from the RCNN. Subsequently, we apply our ranking function (see equation \ref{eq_ranking}) to the pairs of objects, as shown in Figure \ref{fig_pipeline}. A triple with its corresponding bounding boxes is considered correctly detected, if the triple is similar to the ground truth, and if the union of the bounding boxes has at least 50 percent overlap with the union of the ground truth bounding boxes.

\paragraph{Relationship detection:} The second setting, which is also considered in \cite{visual} is relationship detection. It is similar to phrase detection, but with the difference that it is not enough when the union of the bounding boxes is overlapping by at least 50 percent. Instead, both the bounding box of the \textit{subject} and the bounding box of the \textit{object} need at least 50 percent of overlap with their ground truth.

\paragraph{Triple detection:} We add a setting, which we call triple detection, which evaluates only the prediction of the triples. A triple is correct if it corresponds to the ground truth. The position of the predicted bounding boxes is not evaluated.

\paragraph{Predicate detection:} In predicate detection,  it is assumed that \textit{subject} and \textit{object} are given, and only the correct \textit{predicate} between both needs to be predicted. Therefore, we use the ground truth bounding boxes with the respective labels for the objects instead of the bounding boxes derived by the RCNN. This separates the problem of object detection from the problem of predicting relationships.\newline

\noindent For each test image, we create a ranked list of triples. Similar to \cite{visual} we report the recall at the top 100 elements of the ranked list and the recall at top 50. Note, that there are 700000 possible triples, out of which the correct triples need to be ranked on top. 

When training the semantic model, we hold out 5 percent of the nonzero triples as a validation set. We determine the optimal rank for the link prediction methods based on that hold-out set. For the visual model (RCNN and VGG) we use a pretrained model provided by \cite{visual}.

\begin{figure}[t]
\centering
    \includegraphics[width=0.5\textwidth]{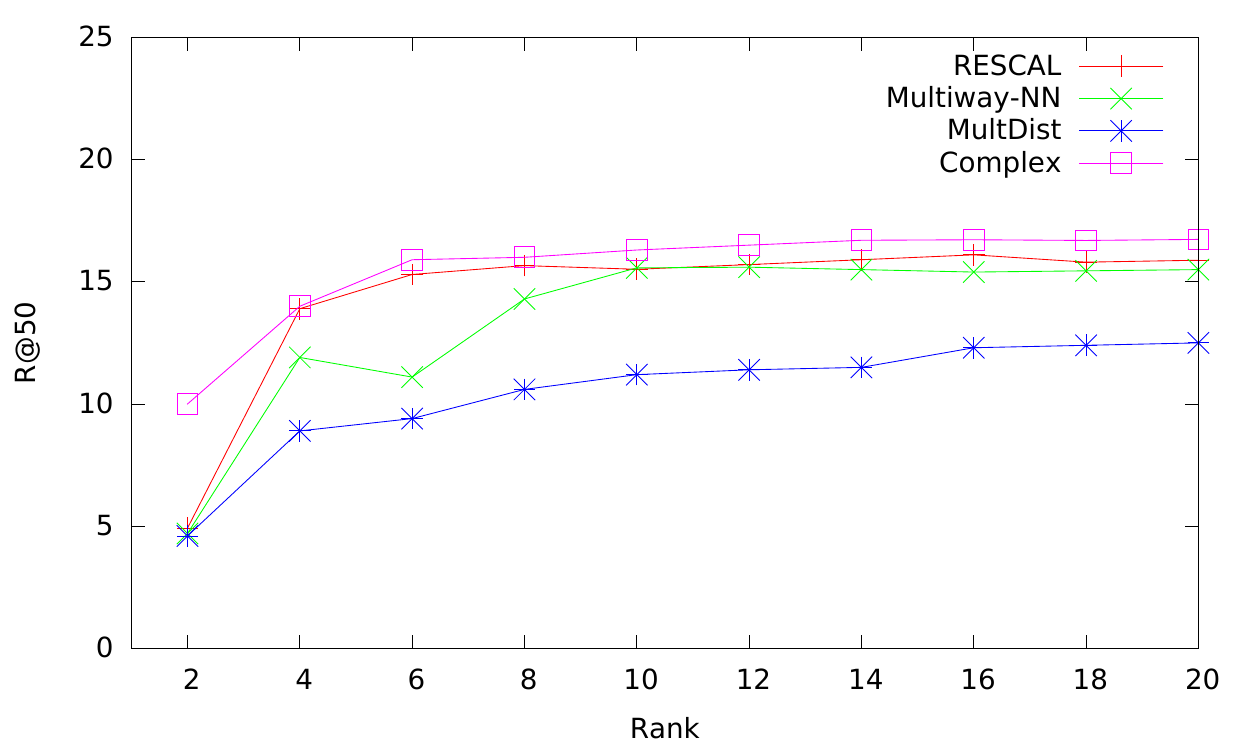}
    \caption{Recall at 50 as a function of the rank}
    \label{fig_rank_normal}
\end{figure}

\subsubsection{Results}
Table \ref{results_main} shows the results for visual relationship detection. The first row shows the results, when only the visual part of the model is applied. This model performs poorly, in all four settings. The full model in the second row adds the language prior to it and also some regularization terms during training, which are described in more detail in \cite{visual}. This drastically improves the results. As expected the recall at top 100 is better than at top 50, however the difference is rather small, which shows that most of the correctly ranked triples are ranked quite high. The results for predicate detection are much better than for the other settings. This shows that one of the main problems in visual relationship detection is the correct prediction of the entities. In the last four rows we report the results of our method, which adds a link prediction model to the visual model. We compare the results for the integration of the four link prediction methods described in section \ref{section_related_work_link_prediction}. We see that with all link prediction methods the model performs constantly better than the state-of-the-art method proposed by \cite{visual}, except for \textit{DistMult}. For \textit{Relationship detection}, which is the most challenging setting, \textit{ComplEx} works best, with a recall of 17.12 and 16.03 for the top 100 and top 50 results respectively. \textit{RESCAL} performs slightly better than the \textit{Multiway Neural Network} in all evaluation settings. For the setting of \textit{Triple Detection} the scores are higher for all methods, as expected, as the overlap of the bounding boxes is not taken into account. However, the relative performance between the methods does not vary much.

Figure \ref{fig_rank_normal} shows the recall at 50 on the test set for our different variants as a function of the rank. We see that the performances of \textit{ComplEx} and \textit{RESCAL} converge relatively quickly to a recall of around 16. The \textit{Multiway Neural Network} converges a bit slower, to a slightly smaller maximum. \textit{DistMult} converges slower and to a much smaller maximum recall of 12.5.

\subsection{Zero-shot Learning}

\begin{figure}[t]
\centering
    \includegraphics[width=0.5\textwidth]{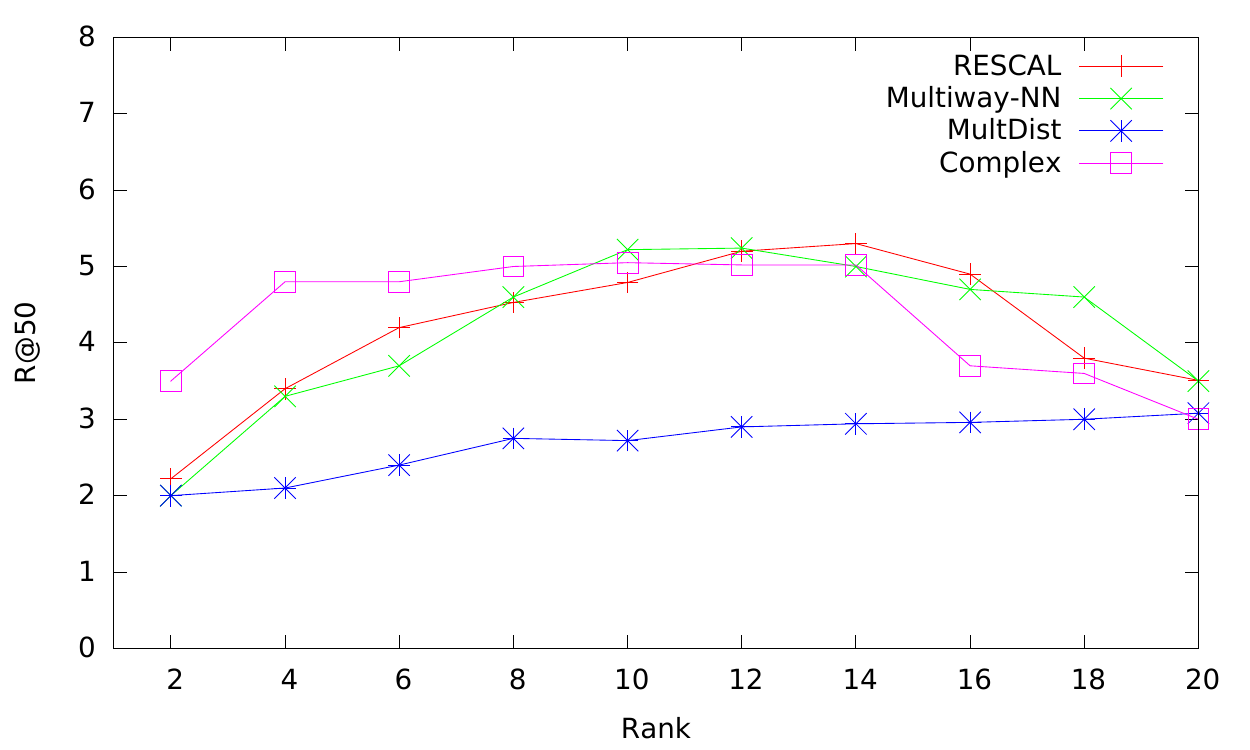}
    \caption{Recall at 50 as a function of the rank for the zero-shot setting.}
    \label{fig_rank_zero}
\end{figure}

\subsubsection{Experimental Setting}

We also include an experimental setting, where we only evaluate on triples, which had not been observed in the training data. This setting reveals the generalization ability of the semantic model. The test set contains 1877 of these triples. We evaluate based on the same settings as in the previous section, however for the recall we only count how many of the unseen triples are retrieved.

\subsubsection{Results}
Table \ref{results_zeroshot} shows the results for the zero-shot experiments. This task is much more difficult, which can be seen by the huge drop in recall. However, also in this experiment, including the semantic model significantly improves the prediction. For the first three settings, the best performing method, which is the \textit{Multiway Neural Network}, almost retrieves twice as many correct triples, as the state-of-the-art model of \cite{visual}. Especially, for the \textit{Predicate Detection}, which assumes the objects and subjects to be given, a relatively high recall of 16.60 can be reached. In the zero-shot setting for \textit{Predicate Detection} even the integration of the worst performing semantic model \textit{DistMult} shows significantly better performance than the state-of-the-art method. These results clearly show that our model is able to infer also new likely triples, which have not been observed in the training data. This is one of the big benefits of the link prediction methods.

Figure \ref{fig_rank_zero} shows the recall at 50 on the zero-shot test set as a function of the rank. As expected, the models start to overfit in the zero-shot setting if the rank is to high. With a limited rank the models have less freedom for explaining the variation in the data; this forces them to focus more on the underlying structure, which improves the generalization property. \textit{ComplEx}, which has more parameters due to the complex valued embeddings, performs best with small ranks and reaches the maximum at a rank of around 8. \textit{Multiway Neural Network} reaches the maximum at a rank of 10 and \textit{RESCAL} at a rank of 14. The highest recall is achieved by \textit{RESCAL} at 5.3.

%-We use pretrained model provided by feifei

\begin{table}[t]
\centering
\caption{Results for the zero shot learning experiments. We report Recall at 50 and 100 for four different validation settings.}
\setlength{\tabcolsep}{0.4em}
{\renewcommand{\arraystretch}{1.4}
\begin{tabular}{|c||c|c||c|c||c|c||c|c|}
\hline
 Task &\multicolumn{2}{c||}{Phrase Det.} &\multicolumn{2}{c||}{Rel. Det.} &\multicolumn{2}{c||}{Predicate Det.} & \multicolumn{2}{c|}{Triple Det.} \\
\hline Evaluation & R@100 & R@50 & R@100 & R@50 & R@100 & R@50 & R@100 & R@50\\
\hline
\hline Lu et al. V \cite{visual} & 1.12 & 0.95 & 0.78 & 0.67 & 3.52 & 3.52 & 1.20 & 1.03 \\ 
\hline Lu et al. full \cite{visual} & 3.75 & 3.36 & 3.52 & 3.13 & 8.45 & 8.45 & 5.39 & 4.79 \\ 
\hline \hline RESCAL & 6.59 & 5.82 & 6.07 & 5.30 & 16.34 & 16.34 & 6.07 & 5.30\\
\hline MultiwayNN & 6.93 & 5.73 & 6.24 & 5.22 & 16.60 & 16.60 & 6.24 & 5.22 \\
\hline ComplEx & 6.50  & 5.73 & 5.82 & 5.05 & 15.74 & 15.74 & 5.82 & 5.05 \\ %rank 100
\hline DistMult & 4.19 & 3.34 & 3.85 & 3.08 & 12.40 & 12.40 & 3.85 & 3.08 \\
\hline 
\end{tabular} 
}
\label{results_zeroshot}
\end{table}

\section{Conclusion}
\label{section_conclusion}

We presented a novel approach for including semantic knowledge into visual relationship detection. We combine a state-of-the-art computer vision procedure with latent variable models for link prediction, in order to enhance the modeling of relationships among visual objects. By including the semantic model, the predictive quality can be enhanced significantly. Especially the prediction of triples, which have not been observed in the training data, can be enhanced through the generalization properties of the semantic link prediction methods. The recall of the best performing link-prediction method in the zero-shot setting is almost twice as high as the state-of-the art method. We proposed a probabilistic framework for integrating both the semantic prior and the computer vision algorithms into a joint model. This paper shows how the interaction of semantic and perceptual models can support each other to derive better predictive accuracies. The developed methods  show great potential also for broader application areas, where both semantic and sensory data is observed. For example, in an industrial setting it might be interesting to model sensor measurements from a plant jointly with a given ontology. The improvement over the state-of-the-art vision model shows that it is not only beneficial to search for better computer vision models; also modeling the semantic part is of central importance. In future work the design of a more expressive ontology might even further improve the results.

%Search (Sparql queries on images!)

\bibliographystyle{splncs03}
\bibliography{references}

\end{document}